\documentclass[10pt,twocolumn,letterpaper]{article}

\usepackage{cvpr}              

\usepackage{graphicx}
\usepackage{amsmath}
\usepackage{amssymb}
\usepackage{booktabs}
\usepackage{makecell,multirow,diagbox}
\usepackage{color}

\definecolor{cvprblue}{rgb}{0.21,0.49,0.74}
\usepackage[pagebackref,breaklinks,colorlinks,allcolors=cvprblue]{hyperref}

\def\paperID{5069} 
\def\confName{CVPR}
\def\confYear{2026}

\title{Cross-Coordinate Correspondence Pruning for Image-to-Point Cloud Registration}

\makeatletter
\def\@author{
    Xin Liu$^{1,*}$, Rong Qin$^{1}$\thanks{Both authors contributed equally to this research.}, Huipeng Lin$^{1}$, Leizhi Shu$^{4}$, Jin Wu$^{5}$, Chi-Man Vong$^{6}$, Liang Lin$^{2,7}$, Jufeng Yang$^{1,2,3}$\thanks{Corresponding author.} \\
    {\small $^1$ VCIP \& TMCC \& DISSec, College of Computer Science, Nankai University, Tianjin, China.} \\
    {\small $^2$ Pengcheng Laboratory, Shenzhen, China.} \\
    {\small $^3$ Nankai International Advanced Research Institute (SHENZHEN·FUTIAN), Shenzhen, China.} \\
    {\small $^4$ AutoCity (Shenzhen) Autonomous Driving Co., Ltd, Shenzhen, China.} \\
    {\small $^5$ School of Intelligent Science and Technology, University of Science and Technology Beijing, Beijing, China.} \\
    {\small $^6$ Department of Computer and Information Science, University of Macau, Macau, China} \\
    {\small $^7$ School of Computer Science and Engineering, Sun Yat-sen University, Guangzhou, China} \\
}

\begin{document}
\maketitle
\begin{abstract}  
%
%
Recent detection-free approaches have shown significant efficacy in image-to-point cloud (I2P) registration by employing a coarse-to-fine matching pipeline.
In the coarse stage, down-sampled image features and voxelized point cloud features are typically fused to establish initial coarse correspondences for subsequent refinement. 
However, existing methods largely overlook the critical role of point cloud density, which fundamentally dictates the quality of coarse correspondences and the final registration results.
Specifically, excessively sparse point clouds lead to an insufficient number of inliers, while overly dense ones often introduce a high outlier ratio.
Consequently, this creates an inherent density trade-off, thereby significantly limiting the registration accuracy of current approaches.
For mitigating this trade-off, we propose a novel Cross-Coordinate Correspondences Pruning (CCP) strategy to acquire sufficient inliers while ensuring a low outlier ratio. 
To minimize interference from inter-modal coordinate discrepancies, we first project cross-coordinate coarse correspondences to the 2D image coordinate system for spatial unification.
Subsequently, a lightweight pruning network is responsible for predicting the inlier confidences, which are used to filter coarse outliers, from coordinate geometric and modal feature dimensions.
%
To maximize inlier recall, we further design a Multi-Density Point Ensemble (MDPE) strategy that consolidates and deduplicates pruned coarse correspondences across varying point cloud densities.
Our method achieves a significant performance improvement, surpassing existing state-of-the-art methods by at least 8.6\% in \textit{Registration Recall} across various benchmarks.
%
%
%
\end{abstract}
\vspace{-4mm}
\section{Introduction}
Image-to-point cloud (I2P) registration serves as a fundamental task across various domains, such as autonomous driving~\cite{chib2023recent}, robot navigation~\cite{wang2023camo}, and simultaneous localization and mapping~\cite{mur2015orb}.
It aims to determine an optimal rigid transformation for accurately aligning the 2D images and 3D point clouds of the same scene, where the accurate establishment of cross-modal correspondences is paramount.
Existing methods tackle this problem in two ways: detection-based and detection-free paradigms~\cite{li20232d3d}.
The former~\cite{feng20192d3d, pham2020lcd, wang2021p2} adopts a detect-then-match strategy to obtain sparse pixel-to-point matches, which suffers from low inlier ratios due to the phased design. 
While the latter~\cite{li20232d3d, kang2024cofii2p, yao2024quantity} directly generates dense correspondences by aligning 2D and 3D patches at a coarse level and subsequently refining them, achieving improved performance.
%
%
%
%
%
%
\begin{figure*}[t]
	\centering
	\includegraphics[width=\linewidth]{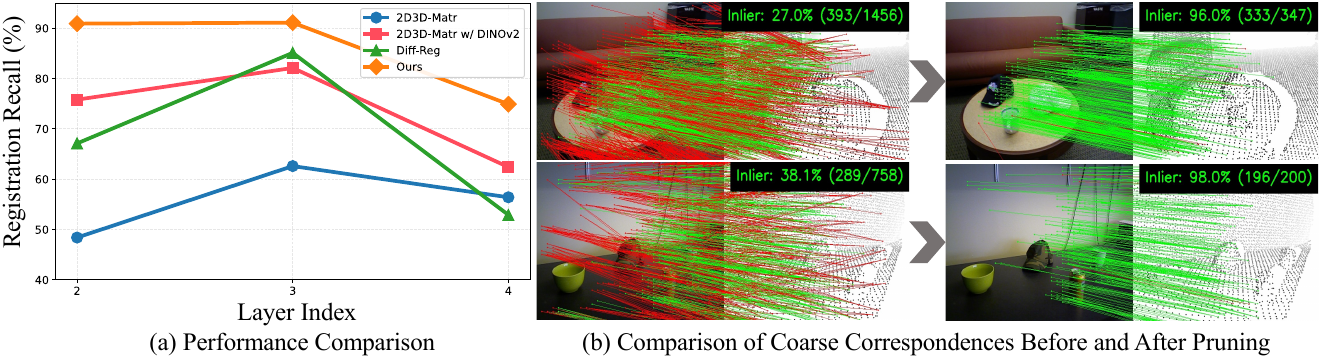}\\  
     \vspace{-0.2cm}
    \caption{(a) Sensitivity analysis of coarse-stage point density. We evaluate the registration performance on the RGB-D Scenes V2~\cite{lai2014unsupervised} by adjusting the density of matchable points across various feature layers. (b) Visualization of coarse correspondences before and after applying the proposed CCP. The inlier ratio is shown top-right.}\label{FIG1}
\end{figure*}
\par
Detection-free methods~\cite{li20232d3d, kang2024cofii2p, wu2024diff, Cheng_2025_ICCV} generally adopt a coarse-to-fine paradigm that involves image downsampling and point cloud voxelization in the coarse stage, followed by cross-modal feature alignment.
As the pioneer, 2D3D-Matr~\cite{li20232d3d} adopts a U-Net structure via attention to learn cross-modality correlations, with coarse matching performed at the deepest feature layers.
Along this line, subsequent methods~\cite{kang2024cofii2p, wu2024diff, Cheng_2025_ICCV} further enhance performance by focusing on cross-modality fusion and alignment.
For example, CA-I2P~\cite{Cheng_2025_ICCV} uses two enhanced modules to improve feature quality and mitigate matching errors.
Diff$^2$I2P~\cite{mu2025diff2i2p} bridges the modality gap by utilizing diffusion prior obtained from the depth-conditioned diffusion model.
%
\par
%
%
%
Crucially, our empirical analysis reveals that the point cloud density employed during the coarse stage serves as a more crucial factor for both coarse correspondence quality and final registration accuracy than the aligning designs.
As illustrated in Fig~\ref{FIG1} (a), all methods demonstrate substantial performance gains when the density of matchable point clouds is appropriately increased at the coarse stages ($e.g.$, at level indices 2 or 3).
%
Nevertheless, as the density of matchable points increases, the proportion of outliers within the coarse correspondences also rises (see ablation study).
%
These excessive outliers introduce substantial noise and interference into the subsequent matching pipeline, ultimately precipitating performance degradation.
%
Consequently, point clouds that are either excessively sparse or overly dense may result in insufficient or unreliable inliers within the coarse correspondences, thereby negatively impacting the refinement process.
To overcome this limitation, it is imperative to develop a strategy that leverages appropriate point cloud density to maximize inlier recall while maintaining a relatively low outlier ratio.
%
We characterize this inherent conflict as the density trade-off.
%
%
%
%
%
%
%
\par
\par
To mitigate this density trade-off, we develop a novel Cross-Coordinate Correspondence Pruning (CCP) strategy that effectively eliminates outliers from coarse correspondences across varying point cloud densities, while retaining a sufficient number of inliers.
Specifically, CCP first establishes initial coarse correspondences between the downsampled image and the voxelized point cloud.
%
To unify the spatial representation, we project 3D point cloud coordinates onto the 2D image plane and fuse them with image coordinates, thereby deriving robust geometric information.
%
%
%
Subsequently, a lightweight MLP architecture functions as the core inference engine of CCP, performing end-to-end learning to predict inlier confidence scores by jointly utilizing coordinate geometry and multi-modal feature representations.
As illustrated in Fig~\ref{FIG1}(b), CCP effectively suppresses coarse outliers and enhances inlier precision based on the confidence scores and a predefined threshold.
%
%
Complementing this pruning mechanism, we introduce a Multi-Density Point Ensemble (MDPE) strategy to maximize inlier recall.
%
MDPE aggregates coarse correspondences derived from varying point cloud densities, which are subsequently filtered via CCP, and then merges and deduplicates the results for optimal downstream refinement.
%
Empirical evaluations show that our method consistently outperforms state-of-the-art methods, improving \textit{Registration Recall} by at least 8.6\% across diverse benchmarks.
\par
Our contributions are threefold:
%
1) We introduce a novel CCP strategy designed to mitigate the inherent density trade-off.
By integrating fused 2D-3D coordinate geometric cues with multi-modal features, CCP performs end-to-end inlier confidence prediction, effectively filtering outliers from coarse correspondences while preserving sufficient inliers
%
%
2) We devise an MDPE strategy to maximize inlier recall.
It aggregates coarse correspondences derived from varying point cloud densities, utilizing a robust merge-and-deduplication process to provide high-fidelity inputs for downstream refinement.
3) Extensive experiments demonstrate that our method outperforms state-of-the-art approaches on multiple benchmarks, yielding significant gains in registration performance.

%
\section{Related Work}
\subsection{Image-to-Point Cloud Registration}
Compared with same-modality registration, I2P registration is more challenging due to substantial cross-modal discrepancies.
Early studies~\cite{feng20192d3d, pham2020lcd, wang2021p2} similarly adopted a detect-then-match pipeline to establish cross-modal correspondences.
For example, 2D3D-Matchnet~\cite{feng20192d3d} extracts SIFT~\cite{lowe2004distinctive} from images and ISS~\cite{sontag1998comments} from point clouds, respectively, to perform matching based on the similarity of descriptors.
Unfortunately, this strategy often results in a low inlier ratio due to the inherent inconsistency between 2D and 3D keypoint detection mechanisms.
More recently, detection-free 2D3D-Matr~\cite{li20232d3d} employs a Transformer-based~\cite{vaswani2017attention} architecture to enable cross-modal interaction and coarse-to-fine matching, leading to notable performance gains.
%
%
Subsequent approaches~\cite{wang2024freereg, kang2024cofii2p, yao2024quantity, wu2024diff, Cheng_2025_ICCV} mainly concentrate on developing strong learning paradigms and network modules to improve accuracy.
FreeReg~\cite{wang2024freereg} unifies heterogeneous modalities through the pre-trained models, which leverages semantically consistent diffusion representations and geometric features, without requiring task-specific training.
Diff-Reg~\cite{wu2024diff} introduces a denoising diffusion process in the space of doubly stochastic matrices to achieve robust correspondence estimation.
%
CA-I2P~\cite{Cheng_2025_ICCV} utilizes the channel adaptive adjustment module to enhance intra-modal features and suppress cross-modal sensitivity.
Nonetheless, existing approaches often neglect a simple yet important factor: the influence of point cloud density during the coarse stage.
We further examine the role of point cloud density in I2P registration and propose an effective pruning mechanism to improve the registration robustness.
%
%
%
%
\vspace{-2mm}
\subsection{Correspondence Pruning}
Correspondence pruning aims at distilling accurate inliers from initial correspondences.
Ranging from classical geometry-based verification~\cite{bian2017gms,lin2017code,ma2019locality} to learning-based paradigms~\cite{yi2018learning,zhang2019learning, bai2021pointdsc,liu2024ncmnet}, these methods have proven highly effective as a post-processing step for robust registration.
For example, LFGC~\cite{yi2018learning} reformulates the problem as a classification task and employs an MLP-based neural network to estimate inlier weights.
NCMNet~\cite{liu2024ncmnet} investigates the synergy among three types of neighbors to enhance the discriminability.
PointDSC~\cite{bai2021pointdsc} further extends this paradigm to point clouds, and explicitly incorporates spatial consistency to enhance feature correlations.
Subsequent approaches~\cite{pais20203dregnet, zhang20233d, yan2025turboreg} have explored diverse learning paradigms or architectural modules to enhance the classification reliability.
Motivated by these insights, we incorporate correspondence pruning into the I2P registration framework to achieve superior registration performance.
%
%
Notably, while existing pruning methods predominantly operate within intra-modal settings, we propose a cross-coordinate projection mechanism to mitigate the structural discrepancy across modalities, which significantly enhances pruning accuracy.
%
%
%
%
%
\par
%
%
%
%
\section{Methodology}
\subsection{Preliminaries}

\textbf{Problem formulation.}
Given the overlapped image $ {\rm \textbf{I}} \in \mathbb{R}^{H \times W \times 3}$ and point cloud $ {\rm \textbf{P}} \in \mathbb{R}^{N \times 3}$, I2P registration aims to estimate the relative rigid transformation ${\rm \textbf{T}} = [\textbf{R} | \textbf{t}]$ for aligning the two modalities, where rotation $\textbf{R} \in SO(3)$ and translation $\textbf{t} \in \mathbb{R}^3$.
Based on the estimated pixel-to-point correspondences $C =\{ {(\mathbf{x}_i,\mathbf{y}_i) | \mathbf{x}_i \in \mathbb{R}^2, \mathbf{y}_i \in \mathbb{R}^3} \}$, the transformation ${\rm \textbf{T}}$ is recovered through the minimization of the 2D projection error:
\begin{equation}
\min_{\mathbf{R},\mathbf{t}} \sum_{(\mathbf{x}_i,\mathbf{y}_i) \in \mathcal{C}} \left| \mathcal{K}(\mathbf{R}\mathbf{y}_i + \mathbf{t}, \mathbf{K}) - \mathbf{x}_i \right|^2 ,
\end{equation}
where $\mathbf{K}$ denotes the camera intrinsic matrix and $\mathcal{K}$ represents the 3D-to-2D projection function.
The optimization problem is efficiently solved via the PnP-RANSAC algorithm~\cite{lepetit2009ep}.
In this process, the accurate estimation of correspondences is of paramount importance and is the focus of this work.
\par

\textbf{Overview of coarse-to-fine matching framework.}
Here, we review the widely applied coarse-to-fine matching framework~\cite{li20232d3d} in I2P registration.
Specifically, in this framework, the image is first processed by a four-stage ResNet~\cite{he2016deep} equipped with FPN~\cite{lin2017feature} to generate multi-resolution 2D features $ {\rm \textbf{F}^I_s} \in \mathbb{R}^{H_s \times W_s  \times C_s}$ and corresponding coordinate matrices $ {\rm \textbf{Q}^I_s} \in \mathbb{R}^{H_s \times W_s  \times 2}$.
Here, $s$ denotes the pyramid level (usually set to 4) and $C_s$ represents the number of channels.
Meanwhile, a four-stage KPFCNN~\cite{thomas2019kpconv} is employed to voxelize the 3D points $ {\rm \textbf{P}^P_s} \in \mathbb{R}^{N_s \times 3}$ and obtain multi-scale features $ {\rm \textbf{F}^P_s} \in \mathbb{R}^{N_s \times C_s}$.
%
%
In the coarse stage, it typically perform patch matching between the lowest resolution of 2D features ($i.e.$, $ {\rm \textbf{F}^I_4}$, $ {\rm \textbf{Q}^I_4}$) and the coarsest scale of 3D features ($i.e.$, $ {\rm \textbf{F}^P_4}$, $ {\rm \textbf{P}^P_4}$).
To enhance the cross-modal correlations, a Transformer-based~\cite{vaswani2017attention} feature fusion module equipped with positional information is used.
Then, the coarse correspondence set $C_{c}$ is established by mutual top-$k$ similarity:
%
%
\begin{equation}
\small
C_{c} = \{ {(\hat{\mathbf{x}}_i,\hat{\mathbf{y}}_i) | \hat{\mathbf{x}}_i \in {\rm \textbf{Q}^I_4}, \hat{ \mathbf{y}}_i \in {\rm \textbf{P}^P_4}} \} = topk( d(\hat{\mathbf{F}}^{\mathrm{I}}_4, \hat{\mathbf{F}}^{\mathrm{P}}_4)  \le \tau_c  ),
\label{eq:coarse_corr}
\end{equation}
\vspace{-4mm}
\begin{equation}
\hat{\mathbf{F}}^{\mathrm{I}}_4, \hat{\mathbf{F}}^{\mathrm{P}}_4 = Att({\rm \textbf{F}^I_4},  {\rm \textbf{Q}^I_4}, {\rm \textbf{F}^P_4}, {\rm \textbf{P}^P_4}),   
\end{equation}
in which $\hat{\mathbf{F}}^{\mathrm{I}}_4$ and $\hat{\mathbf{F}}^{\mathrm{P}}_4$ denote the enhanced 2D and 3D features, respectively.
$d(\cdot,\cdot)$ measures the feature similarity, and $\tau_c$ is the similarity threshold in the coarse stage.
Feature fusion module $Att(\cdot)$ contains both self-attention and cross-attention operations.
Finally, in the fine stage, it further extracts the local dense correspondences in each $(\hat{\mathbf{x}}_i,\hat{\mathbf{y}}_i)$ by leveraging the fine-level 2D and 3D features and mutual top-$k$ selection scheme.
This process can be formulated as:
%
%
\vspace{1mm}
\begin{equation}
\begin{aligned}
C_{f} &= \{ (\mathbf{x}_i,\mathbf{y}_i) \mid \mathbf{x}_i \in \mathbf{Q}^I_1, \mathbf{y}_i \in \mathbf{P}^P_1 \} \\
&= \mathrm{topk}(d(\mathbf{F}^{\mathrm{I}}_1, \mathbf{F}^{\mathrm{P}}_1) \le \tau_f ), \quad \forall (\mathbf{x}_i, \mathbf{y}_i) \in C_c,
\end{aligned}
\label{fine}
\end{equation}
where $\tau_f$ is the similarity threshold in the fine stage.
%
$C_{f}$ is the fine pixel-to-point correspondence set.
%
\begin{figure*}[t]
	\centering
	\includegraphics[width=0.95\linewidth]{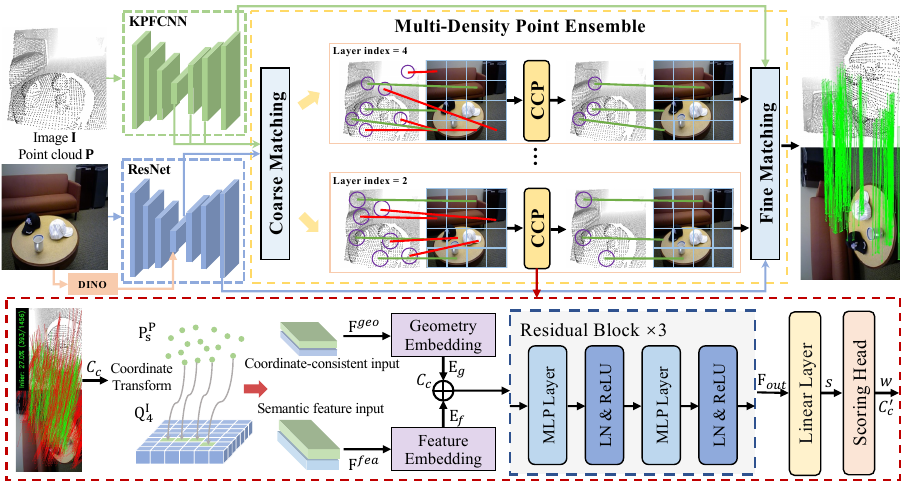}\\  
    \caption{Overview of our proposed pipeline. The framework consists of three core stages~\cite{li20232d3d}: (1) Feature Extraction, where the image $ {\rm \textbf{I}}$ and point cloud $ {\rm \textbf{P}}$ are encoded into multi-scale 2D and 3D features via ResNet~\cite{he2016deep} and KPFCNN~\cite{thomas2019kpconv}, respectively; (2) Coarse Matching, leveraging the proposed CCP and MDPE strategies to maximize the acquisition of robust coarse correspondences; and (3) Fine Matching, which refines the initial matches to produce high-fidelity, fine-grained correspondences.}\label{FIG2}
     \vspace{-0.5cm}
\end{figure*}

\subsection{Limitation of Coarse-to-Fine Matching}\label{limitation}
%
%
Based on the Eq.~\ref{fine}, the effectiveness of the coarse-to-fine matching framework is built upon a high-quality coarse correspondence set,  $i.e.$, $C_{c}$.
A commonly overlooked limitation is that, to meet efficiency requirements, excessive downsampling or voxelization of the input data leads to the loss of local context and structure, thereby undermining the reliability of coarse correspondences.
This problem becomes even more pronounced for unordered point clouds, as voxelization is inherently more prone to irreversible structural loss compared to regular image downsampling.
To mitigate this problem, raising the point cloud density at the coarse stage appears to be a more feasible solution than attempting to increase the image resolution.
Nevertheless, while this manner will increase the number of inliers in coarse correspondences, it unavoidably brings in additional outliers, which introduces more noise and interference for the subsequent fine stage.
This new challenge can be concluded as the density trade-off.

Fundamentally, we attribute this density trade-off to the degraded filtering capability of cross-modal feature similarity, driven by growing spatial overlap of receptive fields as voxel size decreases.
Specifically, let $v$ denote the voxel size, which bounds the spatial distance $d$ between adjacent voxel-sampled 3D candidates $\mathbf{y}_i$ and $\mathbf{y}_j$ ($d \propto v$).
Cross-modal feature extraction relies on a fixed spherical receptive neighborhood $\Omega(\mathbf{y}_i) = \{\mathbf{y} \mid \|\mathbf{y} - \mathbf{y}_i\| \leq r\}$ with search radius $r \gg d$.
We formally quantify the \textit{neighborhood overlap degree} $\lambda(v,\mathbf{y}_i,\mathbf{y}_j)$ between $\mathbf{y}_i$ and $\mathbf{y}_j$ via the standard Jaccard similarity for 3D spatial overlap:
\begin{equation}
\lambda(v,\mathbf{y}_i,\mathbf{y}_j) = \frac{|\Omega(\mathbf{y}_i) \cap \Omega(\mathbf{y}_j)|}{|\Omega(\mathbf{y}_i) \cup \Omega(\mathbf{y}_j)|}.
\label{eq:overlap_definition}
\end{equation}
Via spherical cap volume calculation and first-order Taylor expansion, we derive the closed-form approximation:
%
\begin{equation}
\lambda(v,\mathbf{y}_i,\mathbf{y}_j)  \approx 1 - \frac{3}{2}\frac{d}{r}.
\label{eq:overlap_ratio}
\end{equation}
This derivation shows $\lambda(v,\mathbf{y}_i,\mathbf{y}_j)$ rises sharply with smaller $v$, causing high overlap and feature homogenization.
Thus, the filtering capability $\mathcal{S}(v,\mathbf{y}_i,\mathbf{y}_j)$ of feature similarity is bounded by non-overlapping proportion and degrades with $v$:
\begin{equation}
\mathcal{S}(v,\mathbf{y}_i,\mathbf{y}_j) \propto 1 - \lambda(v,\mathbf{y}_i,\mathbf{y}_j) \approx \frac{3}{2}\frac{d}{r} \propto \frac{v}{r}.
\label{eq:filtering_degradation}
\end{equation}
This formulation reveals the core challenge that smaller voxels preserve fine geometric details but collapse feature discriminative margin, introducing massive outliers into coarse correspondences.
It is thus imperative to propose a novel filtering strategy tailored for coarse correspondences in dense point clouds.
\subsection{Cross-Coordinate Correspondence Pruning}\label{sec3.3}
To alleviate the aforementioned density trade-off, we propose a cross-coordinate correspondence pruning (CCP) strategy, aimed at retaining sufficient inliers while effectively suppressing outliers.
%
%
%
Following the coarse-to-fine framework~\cite{li20232d3d}, we additionally employ DINOv2~\cite{oquab2023dinov2} as the coarse encoder to capture pixel-level similarities~\cite{wu2024diff} and establish an initial correspondence set $C_c$ via Eq.~(\ref{eq:coarse_corr}).
%
%
%
As mentioned in Sec.~\ref{limitation}, the essential challenge of density trade-off arises from the increased overlap of point cloud neighborhoods when the voxel size decreases.
This leads to a decline in the filtering ability of feature similarity.
%
To overcome this challenge, our CCP strategy introduces a learning-based pruning scheme to provide a new filter from the geometric view and adaptively prune outliers.
\par
%
%
As illustrated in Fig.~\ref{FIG2}, CCP takes the set $C_{c}$ as \textbf{input}, and \textbf{outputs} the inlier confidence of all coarse correspondences.
%
The input $C_{c} = (\hat{\mathbf{x}},\hat{\mathbf{y}}) \in \mathbb{R}^{ N_c \times 5} $ comprises $N_c$ correspondences, each associating a 2D image patch coordinate $\hat{\mathbf{x}}_i \in [0,1]^2$ with a corresponding 3D point cloud coordinate $\hat{\mathbf{y}}_i \in \mathbb{R}^{ N_c \times 3}$.
However, the structural discrepancy between heterogeneous modalities makes it difficult for the model to leverage correspondence consistency to distinguish inliers from outliers.
%
Therefore, we first compute the projected 2D coordinate $\tilde{\mathbf{y}}_i$ for each $y_i = (y_i^1, y_i^2, y_i^3)$ through the perspective transformation $\mathcal{P}(\cdot)$:
%
%
\begin{equation}
\tilde{\mathbf{y}}_i = \mathcal{P}(y_i) =  \left( \frac{f_1 \frac{y_i^1}{y_i^3} + z_1}{W-1}, \frac{f_2 \frac{y_i^2}{y_i^3} + z_2}{H-1} \right) ,
\end{equation}
where $f_i$ and $z_i$ represent the focal lengths and principal point offsets, respectively.
%
$W$ and $H$ denote the width and height of the image, respectively.
%
For 3D points projecting outside the visible camera frustum, we map their coordinates to the image center:
\begin{equation}
\hat{\mathbf{y}}^{\prime}_i = 
\begin{cases} 
\tilde{\mathbf{y}}_i, & \text{if } \tilde{\mathbf{y}}_i \in [0, 1]^2 \\
[0.5, 0.5], & \text{otherwise}.
\end{cases}
\end{equation}
Then, to enhance discrimination, we design a dual-branch fusion module to integrate both coordinate geometric and modal feature.
For input correspondences, we construct two distinct input vectors: a coordinate-consistent vector $\mathbf{F}^{geo}  = (\hat{\mathbf{x}},\hat{\mathbf{y}}^{\prime}) \in \mathbb{R}^{ N_c \times 4} $ and a corresponding semantic feature vector $\mathbf{F}^{fea}  = ({\rm \textbf{F}^I_s} ,{\rm \textbf{F}^P_s}) \in \mathbb{R}^{ N_c \times 2C_s} $.
%
Both inputs are independently projected into the geometry embedding $ {\rm \textbf{E}}_g \in \mathbb{R}^{N_c\times d_m}$ and feature embedding ${\rm \textbf{E}}_f \in \mathbb{R}^{N_c\times d_m}$:
\begin{equation}
{\rm \textbf{E}}_g = \sigma(\text{LN}(\mathbf{W}_g \mathbf{F}^{geo} + \mathbf{b}_g)), {\rm \textbf{E}}_f = \sigma(\text{LN}(\mathbf{W}_f \mathbf{F}^{fea} + \mathbf{b}_f)),
\end{equation}
in which $\text{LN}(\cdot)$ stands for Layer Normalization and $\sigma(\cdot)$ denotes the ReLU activation function.
Next, we utilize the lightweight pruning structure $\mathcal{F_\theta}(\cdot)$, which contains three residual blocks based on MLP, to effectively learn their interaction and obtain the fused representation $\mathbf{F}_{out} \in \mathbb{R}^{N_c\times d_m}$:
\begin{equation}
\mathbf{F}_{out} = \mathcal{F_\theta}({\rm \textbf{E}}_g + {\rm \textbf{E}}_f).
\end{equation}
Finally, the refined features are passed through a scoring head to produce the matching logits $s \in \mathbb{R}^{N_c} $ and inlier confidence $w \in \mathbb{R}^{N_c} $: 
\vspace{-1mm}
\begin{equation}
s = \mathcal{F}_l(\mathbf{F}_{out}), w = \text{Sigmoid}(s) \in (0, 1),
\end{equation}
where linear mapping $\mathcal{F}_l(\cdot)$ maps the channel dimension to 1.
The confidence represents the likelihood of a correspondence being an inlier.
Coarse correspondences with inlier confidence scores below the threshold $\epsilon_{p}$ are discarded, effectively suppressing outliers and facilitating the subsequent fine-matching stage.
%
%
\par
%
%
%
\subsection{Multi-Density Point Ensemble}
To retain as many inliers as possible and enhance the robustness of the matching process, we further design a multi-density point ensemble (MDPE) strategy.
Meanwhile, MDPE also consolidates and removes duplicate pruned fine correspondences across different point cloud densities.
During the coarse matching stage, we consistently employ the enhanced lowest-resolution 2D feature $ \hat{\mathbf{F}}^{\mathrm{I}}_4$ and corresponding coordinate matrix $ {\rm \textbf{Q}^I_4}$ to avoid expensive computational overhead.
For the point cloud modality, we note that aggressive voxelization may cause irreversible structural degradation. Hence, we perform voxelization in a conservative multi-density manner to preserve geometric details.
To be specific, we can obtain the multi-density points $\{ \rm \mathbf{P}_{s}^{P}\}_{s=2}^S$ and corresponding improved features $\{\rm \hat{\mathbf{F}}_{s}^{P}\}_{s=2}^S$ using four-stage KPFCNN, where $s$ denotes the index of the layer.
%
%
For each layer index $s$, an initial coarse correspondence set can be established based on Eq.~(\ref{eq:coarse_corr}).
%
However, although this strategy improves inlier recall, it inevitably introduces a larger number of potential outliers.
To mitigate this, we deploy the CCP strategy for optimization.
Then, the pruned multi-density correspondence set $C_{c}^{\prime}$ is formed by aggregating the high-confidence pairs across all scales.
Finally, based on $C_{c}^{\prime}$, MDPE further removes duplicate pixel-to-point correspondences during the fine-matching stage.
The overall correspondence estimation process can be reformulated as:
%
\begin{equation}
\begin{aligned}
C_{c}^{\prime} &= \{ (\hat{\mathbf{x}}_i^{\prime}, \hat{\mathbf{y}}_i^{\prime}) \mid \hat{\mathbf{x}}_i^{\prime} \in \mathbf{Q}^I_4, \hat{\mathbf{y}}_i^{\prime} \in \mathbf{P}_{s}^P, s \in \{2, \dots, S\}\} \\
&= \bigcup_{s=2}^{S} \mathrm{CCP} ( \mathrm{topk}( d(\hat{\mathbf{F}}^I_4, \hat{\mathbf{F}}_{s}^P) \le \tau_c )),
\end{aligned}
\end{equation}
\vspace{-4mm}
%
%
\begin{equation}
\begin{aligned}
C_{f}^{\prime} &= \{  (\mathbf{x}_i^{\prime},\mathbf{y}_i^{\prime}) \mid \mathbf{x}_i^{\prime} \in \mathbf{Q}^I_1,  \mathbf{y}_i^{\prime} \in \mathbf{P}^P_1  \} \\
&= \Phi (\mathrm{topk}( d(\mathbf{F}^{\mathrm{I}}_1, \mathbf{F}^{\mathrm{P}}_1) \le \tau_f ),  \forall (\mathbf{x}_i^{\prime}, \mathbf{y}_i^{\prime}) \in C_{c}^{\prime} ),
\end{aligned}
\end{equation}
%
where $CCP(\cdot)$ and $\Phi(\cdot)$ represents our CCP strategy and de-duplication operation.
%
%
Overall, the coarse-to-fine framework, strengthened by our multi-density CCP, yields a highly filtered and reliable candidate set. 
This optimization substantially reduces both the search space and error propagation in the subsequent fine-grained matching stage, thereby enhancing final registration accuracy.
\par

\begin{table*}[t]
    \centering
    \tiny
    \caption{Quantitative comparison on both the RGB-D Scenes V2~\cite{lai2014unsupervised} and 7Scenes~\cite{glocker2013real} datasets. Best results are highlighted in \textbf{bold}.}
     \vspace{-0.2cm}
    \label{table1}
    \renewcommand\arraystretch{0.8} 
    \setlength{\tabcolsep}{1.9pt} 
   \resizebox{0.95\linewidth}{!}{ \begin{tabular}{l | c c c c  |c |  c c c c c c c  |c}
        \toprule
        Datasets & \multicolumn{5}{c}{RGB-D Scenes V2} & \multicolumn{8}{c}{7Scenes} \\
        \cmidrule(r){2-6} \cmidrule(l){7-14}
        Classes & S-11 & S-12 & S-13 & S-14 & Mean & Che. & Fire & Hea. & Off. & Pum. & Kit. & Sta. & Mean \\
        \midrule
        Depth& 1.74 & 1.66 & 1.18 & 1.39 & 1.49 & 1.78 & 1.55 & 0.80 & 2.03 & 2.25 & 2.13 & 1.84 & 1.77 \\
        \midrule
        Methods & \multicolumn{13}{c}{\textit{Inlier Ratio (\%) $\uparrow$}} \\
        \midrule
        P2-Net~\cite{wang2021p2} & 9.7 & 12.8 & 17.0 & 9.3 & 12.2 & 55.2 & 46.7 & 13.0 & 36.2 & 32.0 & 32.8 & 5.8 & 31.7 \\
        Pre-2D3D~\cite{huang2021predator} & 17.7 & 19.4 & 17.2 & 8.4 & 15.7 & 34.7 & 33.8 & 16.6 & 25.9 & 23.1 & 22.2 & 7.5 & 23.4 \\
        2D3D-Matr~\cite{li20232d3d}& 32.8 & 34.4 & 39.2 & 23.3 & 32.4 & 72.1 & 66.0 & 31.3 & 60.7 & 50.2 & 52.5 & 18.1 & 50.1 \\
        Diff-Reg~\cite{wu2024diff}& 47.2 & 48.7 & 32.9 & 22.4 & 37.8 & 57.2 & 49.8 & 42.1 & 49.7 & 38.0 & 41.4 & 14.5 & 41.8 \\
        CA-I2P~\cite{Cheng_2025_ICCV} & 38.6 & 40.6 & 38.9 & 24.0 & 35.5 & 73.6 & 66.4 & 34.5 & 62.4 & 52.1 & 52.8 & 19.1 & 51.6 \\
        Diff$^2$I2P~\cite{mu2025diff2i2p} & - & - & - & - & 36.8 & \textbf{74.1} & \textbf{68.8} & 39.2 & \textbf{65.6} & \textbf{52.1} & \textbf{54.2} & 18.1 & 53.2 \\
        Ours & \textbf{56.6} & \textbf{58.9} & \textbf{63.9} & \textbf{51.3} & \textbf{57.7} & 70.4 & 64.3 & \textbf{66.0} & 62.1 & 51.3 & 54.1 & \textbf{35.0} & \textbf{57.7} \\
        \midrule
        Methods & \multicolumn{13}{c}{\textit{Feature Matching Recall (\%) $\uparrow$}} \\
        \midrule
        P2-Net~\cite{wang2021p2} & 48.6 & 65.7 & 82.5 & 41.6 & 59.6 & \textbf{100.0} & 99.3 & 58.9 & 99.1 & 87.2 & 92.2 & 16.2 & 79.0 \\
        Pre-2D3D~\cite{huang2021predator}& 86.1 & 89.2 & 63.9 & 24.3 & 65.9 & 91.3 & 95.1 & 76.7 & 88.6 & 79.2 & 80.6 & 31.1 & 77.5 \\
        2D3D-Matr~\cite{li20232d3d} & 98.6 & 98.0 & 88.7 & 77.9 & 90.8 & \textbf{100.0} & 99.6 & 98.6 & \textbf{100.0} & 92.4 & 95.9 & 58.1 & 92.1 \\
        Diff-Reg~\cite{wu2024diff}& \textbf{100.0} & \textbf{100.0} & 88.7 & 77.0 & 91.4 & \textbf{100.0} & 98.0 & 98.6 & \textbf{100.0} & 86.5 & 96.9 & 47.3 & 89.6 \\
        CA-I2P~\cite{Cheng_2025_ICCV} & \textbf{100.0} & \textbf{100.0} & 91.8 & 82.7 & 93.6 & \textbf{100.0} & \textbf{100.0} & 98.6 & \textbf{100.0} & 92.0 & 95.5 & 60.8 & 92.4 \\
        Diff$^2$I2P~\cite{mu2025diff2i2p} & - & - & - & - & 77.1 & \textbf{100.0} & \textbf{100.0} & \textbf{100.0} & \textbf{100.0} & 93.4 & 96.2 & 55.4 & 92.2 \\
        Ours & \textbf{100.0} & \textbf{100.0} & \textbf{100.0} & \textbf{99.1} & \textbf{99.8} & \textbf{100.0} & \textbf{100.0} & 98.6 & 99.8 & \textbf{96.2} & \textbf{99.6} & \textbf{78.4} & \textbf{96.1} \\
        \midrule
        Methods & \multicolumn{13}{c}{\textit{Registration Recall (\%) $\uparrow$}} \\
        \midrule
        P2-Net~\cite{wang2021p2} & 40.3 & 40.2 & 41.2 & 31.9 & 38.4 & 96.9 & 86.5 & 20.5 & 91.7 & 75.3 & 85.2 & 4.1 & 65.7 \\
        Pre-2D3D~\cite{huang2021predator} & 44.4 & 41.2 & 21.6 & 13.7 & 30.2 & 69.6 & 60.7 & 17.8 & 62.9 & 56.2 & 62.6 & 9.5 & 48.5 \\
        2D3D-Matr~\cite{li20232d3d} & 63.9 & 53.9 & 58.8 & 49.1 & 56.4 & 96.9 & 90.7 & 52.1 & 95.5 & 80.9 & 86.1 & 28.4 & 75.8 \\
        Diff-Reg~\cite{wu2024diff}& 95.2 & 95.1 & 85.6 & 64.6 & 85.1 & 99.7 & 90.9 & 59.0 & 96.7 & 79.9 & 90.9 & 25.8 & 77.5 \\
        CA-I2P~\cite{Cheng_2025_ICCV} & 68.1 & 73.5 & 63.9 & 47.8 & 63.3 & 99.0 & 90.7 & 68.5 & 96.2 & 83.0 & 88.1 & 31.1 & 79.5 \\
        Diff$^2$I2P~\cite{mu2025diff2i2p} & - & - & - & - & 60.5 & 99.0 & 95.6 & 74.0 & 98.9 & 86.8 & 90.2 & 36.5 & 83.0 \\
        Ours & \textbf{98.6} & \textbf{98.0} & \textbf{97.9} & \textbf{83.2} & \textbf{94.4} & \textbf{100.0} & \textbf{97.4} & \textbf{98.6} & \textbf{99.3} & \textbf{86.5} & \textbf{97.2} & \textbf{62.2} & \textbf{91.6} \\
        \bottomrule
    \end{tabular}
    }
     \vspace{-0.3cm}
\end{table*}

\subsection{Loss Functions}
Following~\cite{li20232d3d}, a scaled circle loss~\cite{sun2020circle} $\mathcal{L}_{c}$ at different scales is applied to supervise the coarse-level features, and a standard circle loss $\mathcal{L}_{f}$ is used to optimize dense fine-level image and point cloud features.
Meanwhile, we utilize binary cross-entropy loss $\mathcal{L}_{p}$ to supervise our pruning structure.
%
%
The total loss is calculated as a weighted combination of the three loss terms: $\mathcal{L}_{all} = \mathcal{L}_{c} + \mathcal{L}_{f} + \beta \mathcal{L}_{p}$, where $\beta$ is a weight.
%
%
%
%
%

\section{Experiments}
%
%

\subsection{Datasets and Metrics.}
\textbf{Datasets.}
Following~\cite{li20232d3d}, we conducted extensive experiments and ablation studies to evaluate our method on two widely used indoor benchmarks: RGB-D Scenes V2~\cite{lai2014unsupervised} and 7-Scenes~\cite{glocker2013real}
%
The former comprises 11,427 RGB-D frames captured across 14 indoor scenes, following the standard split where scenes 1 to 8 serve as the training set and scenes 11 to 14 are used for testing. The latter follows its official sequence partitioning protocol, containing 2,304 test pairs, 4,048 training pairs, and 1,011 validation pairs
%
%
\par
\textbf{Metrics.}
%
We evaluate I2P registration performance using three primary metrics:
(1) \textit{Inlier Ratio (IR):} The fraction of fine-level pixel-to-point correspondences with a 3D Euclidean distance below 5 cm.
(2) \textit{Feature Matching Recall (FMR):} The proportion of pairs yielding an IR greater than 0.1.
%
%
(3) \textit{Registration Recall (RR):} This core metric quantifies the proportion of test pairs for which the recovered transformation yields an RMSE lower than 0.1 m.
\begin{figure*}[ht!]
	\centering
	\includegraphics[width=0.9\textwidth]{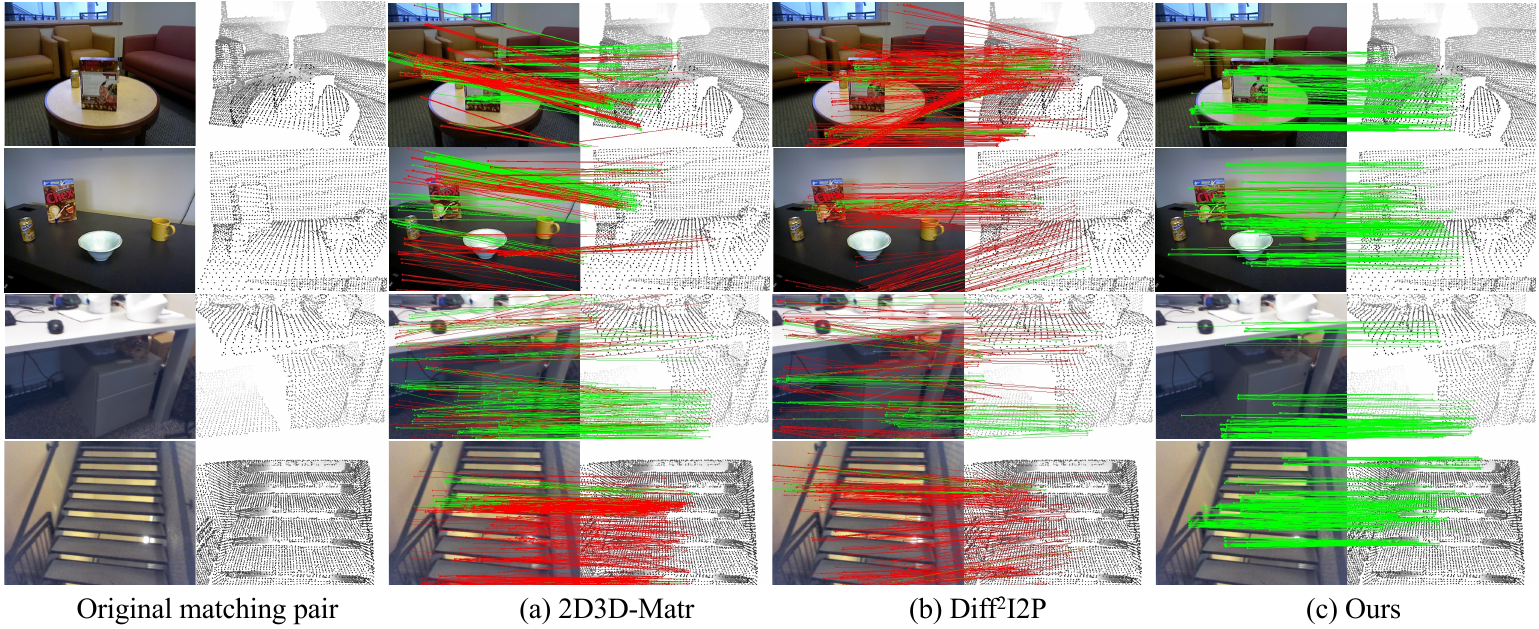}\\
         \vspace{-0.2cm}
   \caption{Qualitative visualization of the top 500 estimated fine correspondences on the RGB-D Scenes V2~\cite{lai2014unsupervised} (top two rows) and 7Scenes datasets. \textcolor{green}{Green} lines denote inliers, while \textcolor{red}{red} lines indicate outliers retained by the networks, respectively.}\label{FIG3}
\end{figure*}
\subsection{Implementation Details.}
%
We build our entire framework with PyTorch, and all training experiments are conducted on an NVIDIA RTX 3090 GPU.
%
For the point cloud backbone, we adopt a four-layer KPFCNN~\cite{thomas2019kpconv} architecture with the initial voxel size of 2.5 cm, which is progressively doubled at each stage to extract multi-density 3D features.
As the 2D feature extractor, we employ a four-layer ResNet~\cite{he2016deep} with a kernel size of 7 to capture multi-resolution 2D features.
The input image is first cropped to $476 \times 630$ and subsequently downsampled to $34 \times 45$ at the coarse stage to ensure compatibility with DINOv2~\cite{oquab2023dinov2}.
The similarity threshold $\tau_c$ and $\tau_f$ are set to 0.95 and 0.75.
In CCP, we adopt three residual blocks with channel dimension $d_m = 256$ as the core component.
The pruning threshold $\epsilon_{p}$ is set to 0.6, and the loss weight $\beta$ is set to 0.5.
In MDPE, we leverage the second- and third-level point cloud features ($i.e.$, $s = 2,3$) to construct the multi-density points.
We adopt the Adam optimizer to train our network for 30 epochs with a batch size of 1. The learning rate is initialized as $1e-4$, and decays by a factor of 0.05 at each training epoch.

\subsection{Comparative Results}
%
%
\par
\begin{table}[t!]
	\centering
	\normalsize
    \small
	\renewcommand\arraystretch{0.95}
    \setlength{\tabcolsep}{1.2pt} 
	\caption{Quantitative comparative results of cross-generalization capabilities across various methods on both the RGB-D Scenes V2~\cite{lai2014unsupervised} and 7Scenes~\cite{glocker2013real} datasets.}\label{Table2}
    \begin{tabular}
			{l | c c c | c c c}
			\toprule
            Datasets & \multicolumn{3}{c|}{RGB-D Scenes V2} & \multicolumn{3}{c}{7Scenes} \\
            \midrule
            Methods & \textit{RR} & \textit{FMR} & \textit{IR} & \textit{RR} & \textit{FMR} & \textit{IR} \\
            \midrule
            2D3D-Matr & 16.3 & 34.9 & 8.5 & 0.1 & 3.4 & 2.3 \\ 
            2D3D-Matr(DINOv2) & 25.8 & 56.8 & 15.9 & 2.0 & 17.5 & 5.3 \\ 
            Diff-Reg & 17.6 & 44.1 & 12.3 & 1.0 & 1.7 & 1.6 \\
            Diff$^2$I2P & 11.9 & 0.5 & 10.3 & 5.4 & 33.7 & 8.0 \\
            \midrule
            Ours & \textbf{54.4} & \textbf{89.4} & \textbf{35.1} & \textbf{33.6} & \textbf{82.5} & \textbf{50.6} \\
            \bottomrule
	\end{tabular}
\end{table}
\textbf{Evaluations on RGB-D Scenes V2.}
As illustrated in the left half of Table~\ref{table1}, our method attains the most significant performance gains compared to the baselines across three key evaluation metrics.
Specifically, in terms of \textit{IR}, it achieves a mean score of 57.7\%, representing a significant margin over Diff-Reg and Diff$^2$I2P.
Regarding \textit{FMR}, our method maintains a near-perfect mean accuracy of 99.8\%, consistently outperforming CA-I2P (93.6\%) and showing remarkable stability across all scene categories.
Finally, benefiting from the accurate fine correspondences, it achieves a superior average  \textit{RR} of 94.4\%, representing a significant improvement over the 85.1\% reported by Diff-Reg.
%
Notably, even in the most challenging S-14 scene where baseline performances typically degrade, our approach retains high accuracy.

\par

\begin{figure}[t]
	\centering
	\includegraphics[width=0.95\linewidth]{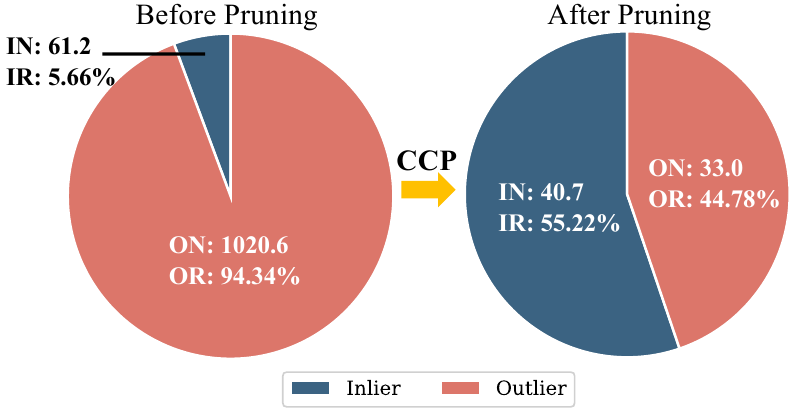}\\
	\caption{The mean of inlier/outlier number (IN/ON) and ratio (IR/OR) before and after pruning on the 7Scenes~\cite{glocker2013real}.}\label{fig:qualitative}
     \vspace{-0.3cm}
\end{figure}

\begin{table*}[ht!]
	\centering
	\footnotesize
	\renewcommand\arraystretch{0.9}
	\caption{Ablation studies regarding the gains of key components in the proposed CCP and MDPE strategies.}\label{xiaorong_main} 
     \vspace{-0.2cm}
    \setlength{\tabcolsep}{5.0pt}
	\resizebox{0.9\linewidth}{!}{\begin{tabular}
			{l  | c c c | c c c}
                \toprule
                Datasets&\multicolumn{3}{c|}{RGB-D Scenes V2} &\multicolumn{3}{c}{7Scenes}\\
                \hline
                Methods & \textit{RR} &  \textit{FMR} & \textit{IR} &  \textit{RR} &  \textit{FMR} & \textit{IR}  \\
                \hline
                (a.1) Ours w/ 3D CT, GE,  FE, layer index s = 2, 3 &94.4 & 99.8 & 57.7 & 91.6 &96.1 & 57.7 \\
                (a.2) Ours (w/o DINOv2) w/ 3D CT, GE,  FE, layer index s = 2, 3 & 91.4& 97.9& 55.4 & 89.7 &95.9&59.2 \\
                \hline
                (b.1) 2D3D-Matr (raw) w/ layer index s = 4   & 56.4  & 90.8  & 32.4  & 75.8 & 92.1  & 50.1   \\
                (b.2) 2D3D-Matr (DINOv2) w/ layer index s = 4   & 62.4  & 94.0  & 37.3  & 79.0 & 91.6 & 51.8  \\
                (b.3) 2D3D-Matr (DINOv2) w/ layer index s = 3  &82.1 & 95.2 & 47.3 & 83.7& 91.4& 52.8 \\
                (b.4) 2D3D-Matr (DINOv2) w/ layer index s = 2   & 75.8& 92.6& 34.2 & 83.8 & 90.7 & 63.4  \\      
                \hline
                \multicolumn{7}{c}{Cross-Coordinate Correspondence Pruning} \\
                \hline 
                (c.1) CCP w/ 2D Coordinate Transform (2D CT)   & 86.7 & 96.1  & 42.7 & 87.2 & 91.6 & 54.8 \\ 
                (c.2) CCP w/o 3D Coordinate Transform (3D CT)  & 86.7 & 95.4  &46.1   & 86.8  & 91.2 & 53.2 \\     
                (c.3) CCP w/o Geometry Embedding (GE)  & 86.1 & 96.2  &40.3  & 86.4  & 91.7 & 53.9 \\
                (c.4) CCP w/o Feature Embedding (FE) & 92.5  & 98.8  & 53.0  & 90.3  & 94.4 & 54.2 \\        
                \hline
                \multicolumn{7}{c}{Multi-Density Point Ensemble} \\
                \hline   
                (d.1) MDPE w/ layer index s = 2, 3, 4  & 92.0  & 99.9  & 55.3  & 90.3  & 97.0 & 56.2\\
                (d.2) MDPE w/ layer index s = 4  & 74.9 &  97.2 & 46.8 & 85.7  & 95.4  & 52.7 \\   
                (d.3) MDPE w/ layer index s = 3  & 91.1  & 99.7 & 55.9  & 89.9  & 96.8 & 59.5 \\ 
                (d.4) MDPE w/ layer index s = 2 & 90.9 & 99.9  & 62.6 & 90.4  & 95.1  &  63.7  \\ 
                \bottomrule
	\end{tabular}}
\end{table*}
\begin{figure*}[ht!]
	\centering
 	\includegraphics[width=0.90\linewidth]{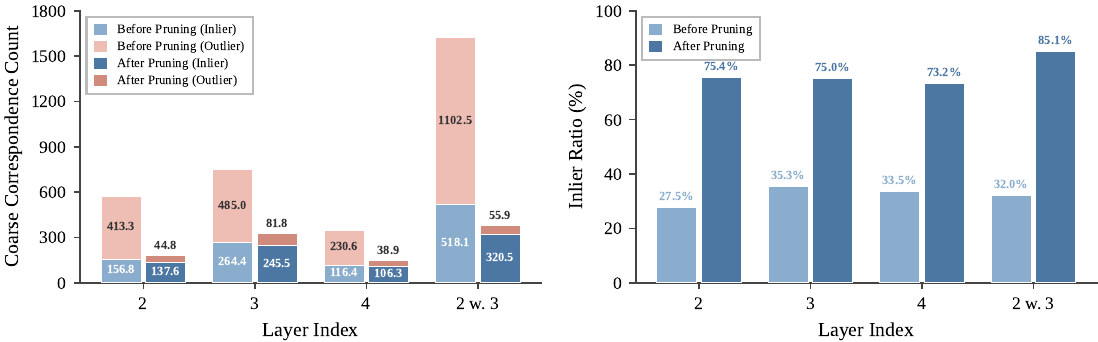}\\  
    \vspace{-0.3cm}
	\caption{Ablation analysis of feature layer indices in KPFCNN~\cite{thomas2019kpconv} on the RGB-D Scenes V2~\cite{lai2014unsupervised} dataset. Comparison of the mean number of coarse correspondences and the inlier ratio at the coarse stage before and after applying our CCP.}\label{xiaorong1}
    \vspace{-0.3cm}
\end{figure*}

\textbf{Evaluations on 7Scenes.}
The results on the 7Scenes dataset further validate the superior robustness of our proposed method in more complex and diverse indoor environments as shown in the right half of Table~\ref{table1}.
Compared to the RGB-D Scenes V2, the 7Scenes dataset poses significant challenges due to its varying scene scales 
and intricate geometric structures.
Despite these difficulties, our approach consistently achieves state-of-the-art results across all seven scenes.
Specifically, for \textit{IR}, it attains a mean of 57.7\%, notably excelling in the highly challenging \textbf{Heads} and \textbf{Stairs} scenes where baselines struggle to maintain precision. 
%
%
%
This consistent performance extends to \textit{FMR}, where we achieve a mean of 96.1\%, underscoring the robustness of our approach across diverse categories.
Most importantly, our method realizes a significant leap in \textit{RR}, reaching a mean of 91.6\%—an 8.6\% absolute improvement over the best baseline Diff$^2$I2P.
%
%
\par
\textbf{Visualization.}
Qualitative results in Fig.~\ref{FIG3} show that our method achieves superior correspondence accuracy with significantly fewer outliers compared to state-of-the-art baselines. 
This visual evidence directly corroborates the substantial quantitative performance gains across both benchmarks reported in Table~\ref{table1}.

\par
\textbf{Generalization.}
To further validate the generalization capability of our framework, we conducted cross-dataset experiments by evaluating models reciprocally on the RGB-D V2 and 7Scenes datasets. 
As illustrated in Table~\ref{Table2}, existing methods often struggle with cross-domain generalization.
In contrast, our pruning strategy leverages inherent geometric constraints, which generally exhibit strong generalization capability~\cite{bai2021pointdsc, qin2023geotransformer}. 
Consequently, our method effectively filters out the majority of outliers, thereby enabling robust registration performance.
Our method maintains robust registration performance across all metrics compared to existing baselines.
We further quantify the distribution of inliers and outliers before and after our CCP as illustrated in Fig.~\ref{fig:qualitative}.
Notably, the inlier ratio improves substantially from 5.66\% to 55.22\% after pruning.
%

%
%
%
%
\par
\subsection{Ablation Study}
%
\par
%
%
%
%

%
\par
\textbf{The CCP strategy.}
Compared with density-varied baselines (b.2–b.4), our CCP pruning strategy (d.2–d.4) substantially improves registration accuracy, verifying CCP’s effectiveness.
Proper coordinate transformation is critical for optimal CCP performance. Removing the 3D transform (c.2, Sec. 3.3) reduces RR by 7.3\% on RGB-D V2 and 4.8\% on 7Scenes, confirming that unifying cross-view geometric coordinates into a shared metric space is necessary to mitigate structural discrepancy. Conversely, projecting image coordinates to 3D space as CCP input (c.1) also severely degrades performance, caused by inherent depth ambiguity that undermines geometric reliability for pruning.
Furthermore, geometric embedding plays a more dominant role than feature embedding (c.3, c.4), indicating correspondence pruning is primarily driven by geometric structural consistency, consistent with prior 2D/3D studies~\cite{yi2018learning,bai2021pointdsc,liu2024ncmnet}.

\par
\textbf{The MDPE strategy}. 
%
%
Point cloud density is critical to registration performance. Extracting features from different KPFCNN layers (varying point densities) in the coarse stage causes notable performance gaps, consistent across both baselines (b.2–b.4) and our approach (d.2–d.4). Excessive voxelization ($e.g.$, s = 4) severely degrades registration, underscoring the need for optimal density configuration. To maximize inlier recall and robustness, our MDPE aggregates coarse correspondences across multiple densities. Combined with our CCP strategy (s = 2, 3), it delivers the highest coarse inlier ratio and best overall performance (a.1).
Furthermore, adding DINOv2 to the 2D3D-Matr baseline (b.1–b.4) yields smaller gains than density adjustment — our core focus. Removing DINOv2 from our full pipeline (a.2) only causes marginal performance drop, empirically confirming that primary improvements come from our proposed strategies.

\par
\textbf{Numerical analysis}. Finally, Fig.~\ref{xiaorong1} further illustrates the quantitative impact of CCP pruning on both mean coarse correspondence count and inlier ratio.
%
%
It is evident that the inlier ratio within the initial coarse correspondences, derived via mutual top-$k$ similarity, before pruning is relatively low.
The proposed CCP effectively removes the majority of outliers while preserving most inliers.
Meanwhile, when combined with MDPE (column 4), both the cardinality and the quality of the inliers are substantially improved.
Consequently, our approach achieves significant performance gains over the baselines.
%
%
\par
\par
\par
%
%
%
%
%
%
\vspace{-2mm}
\section{Conclusion}
Motivated by the inherent limitation of the coarse-to-fine matching pipeline in I2P registration, we propose two core strategies to alleviate the density trade-off.
Specifically, the proposed CCP significantly improves the quality of coarse correspondences by estimating reliable inlier confidences via a lightweight pruning network.
Concurrently, by projecting geometric embedding into the 2D image coordinate space, we effectively mitigate the modal discrepancies inherent in cross-modal registration.
Complementary to this, MDPE optimizes inlier recall through the aggregation and deduplication of fine correspondences across diverse point cloud densities.
The synergistic integration of CCP and MDPE enables our framework to generate high-fidelity coarse correspondences, thereby establishing a robust foundation for subsequent fine-grained refinement.
%
Extensive experiments on challenging benchmarks demonstrate the superior robustness and generalization capability of the proposed approach.
\vspace{-2mm}

{
    \small
    \bibliographystyle{ieeenat_fullname}
    \bibliography{main}
}

\end{document}